# Fuzzy logic and probability

**Petr Hájek**
Institute of Computer Science (ICS)
Academy of Sciences
182 07 Prague, Czech Republic

**Lluís Godo**          **Francesc Esteva**
Institut d'Investigació en Intel.ligencia Artificial (IIIA)
Spanish Council for Scientific Research (CSIC)
08193 Bellaterra, Spain

## Abstract

In this paper we deal with a new approach to probabilistic reasoning in a logical framework. Nearly almost all logics of probability that have been proposed in the literature are based on classical two-valued logic. After making clear the differences between fuzzy logic and probability theory, here we propose a *fuzzy* logic of probability for which completeness results (in a probabilistic sense) are provided. The main idea behind this approach is that probability values of crisp propositions can be understood as truth-values of some suitable fuzzy propositions associated to the crisp ones. Moreover, suggestions and examples of how to extend the formalism to cope with conditional probabilities and with other uncertainty formalisms are also provided.

## 1 Introduction

Discussions about the relation between fuzzy logic and probability are still numerous and sometimes rather controversial. In particular, using fuzzy logic to reason in a probabilistic way may be a priori considered as a "dangerous mixture" of both formalisms. In this sense, the aim of this paper is twofold. First to stress the differences between fuzzy logic and probability theory, making clear that they are different formalisms addressing different problems and using different techniques. Second to show how it is possible to consistently use them together by proposing a new and meaningful approach to probabilistic reasoning based on fuzzy logic. The topic of relating probability and logic is not by far new. A number of logics of probability have been proposed in the literature, as those in [Scott and Kraus, 1966; Hajek and Havranek, 1978; Gaifman and Snir, 1982; Nilsson, 1986; Bacchus, 1990; Halpern, 1989; Wilson and Moral, 1994]. But all of them except for [Hajek and Havranek, 1978] are based on classical two-valued logic. Here we propose a propositional *fuzzy* logic of probability for which complete-ness results are provided. The main idea behind this approach is that probability values of crisp propositions can be understood as truth-values of some suitable fuzzy propositions associated to the crisp ones.

Before going to the technical details in next sections, and in order to avoid misunderstandings, we start by addressing and clarifying the main notions involved in this paper.

## Main difference between fuzzy logic and probability theory

In our opinion any serious discussion on the relation between fuzzy logic and probability must start by making clear the basic differences. Admitting some simplification, we consider that fuzzy logic is a logic of vague, imprecise notions and propositions, propositions that may be more or less true. Fuzzy logic is then a logic of partial degrees of truth. On the contrary, probability deals with crisp notions and propositions, propositions that are either true or false; the probability of a proposition is the degree of belief on the truth of that proposition. If we want to consider both as uncertainty degrees we have to stress that they represent very different sorts of uncertainty (Zimmermann calls them linguistic and stochastic uncertainty, respectively). If we prefer to reserve the word "uncertainty" to refer to degrees of belief, then clearly fuzzy logic does not deal with uncertainty at all. The main difference lies in the fact that degrees of belief are not extensional (truth-functional), e.g. the probability of $p \wedge q$ is not a function of the probability of $p$ and the probability of $q$, whereas degrees of truth of vague notions admit truth-functional approaches (although they are not bound to them). Formally speaking, fuzzy logic behaves as a many-valued logic, whereas probability theory can be related to a kind of two-valued modal logic (cf. e.g. [Hajek, 1993] or [Hajek, 1994] for more details, also [Klir and Folger, 1988]). Thus, fuzzy logic is *not* a "poor man's probability theory", as some people claim.

## Comparing fuzzy logic and probability

Nevertheless, relationships between fuzzy logic and probability theory have been studied. They have not only been compared but also combined. First of all, we refer to [Zadeh, 1986]; even if the title of Zadeh's paper ends with the words "a negative view", he is rather positive in combining fuzziness and probability by suggesting a definition of the probability of a fuzzy proposition. Another important paper is [Dubois and Prade, 1993], in which the authors extensively survey the literature concerning the relationship between fuzzy sets and probability theories; again, besides pointing out the gaps between them, the authors build bridges between both theories, stressing in this sense the importance of possibility theory. Our paper is an attempt to contribute further to this bridge building.

## Probability is ground on classical equivalence

We restrict ourselves to propositional calculus; i.e. formulas are built from propositional variables and connectives (negation $\neg$, implication $\rightarrow$ and possibly others. We shall consider only calculi in which other connectives are definable from negation and implication.

Formulas can be endowed with various semantics, among them the *classical* (boolean, two-valued): there are just two truth values 0 and 1, each evaluation $e$ of propositional variables by zeros and ones extends uniquely to an evaluation of all formulas using classical truth tables. Two formulas $\varphi, \psi$ are *classically equivalent* if $e(\varphi) = e(\psi)$ for any boolean evaluation $e$. The second semantics of our interest is that of *Lukasiewicz's infinite-valued calculus*: truth values are real numbers from the unit interval $[0, 1]$, and truth functions are

$$\begin{aligned} \neg x &= 1 - x \\ x \rightarrow y &= min(1, 1 - x + y) \end{aligned}$$

(we take the freedom of denoting a truth function by the same symbol as its corresponding connective). Under this second semantics, $\varphi, \psi$ are *L-equivalent* if $e(\varphi) = e(\psi)$ for any real-valued Lukasiewicz evaluation $e$. Neglecting the difference between classical and many-valued equivalence of formulas has been the source of known misunderstandings concerning fuzzy logic. Clearly, if two formulas are L-equivalent then they are classically equivalent, but the converse does not hold.

On the other hand, a (finitely additive) *probability* on formulas is a mapping $P$ assigning to each formula $\varphi$ a real number $P(\varphi)$ in $[0, 1]$ preserving classical equivalence (i.e. if $\varphi, \psi$ are classically equivalent then $P(\varphi) = P(\psi)$) and satisfying the well known conditions: $P(true) = 1, P(false) = 0$, and if $\varphi \wedge \psi$ is classically equivalent to $false$ then $P(\varphi \vee \psi) = P(\varphi) + P(\psi)$. Here $true$ is a formula identically true, e.g. $p \rightarrow p$, $false$ is $\neg true$, $\varphi \vee \psi$ is $(\varphi \rightarrow \psi) \rightarrow \psi$ (this is a possible definition of disjunction from implication) and $\varphi \wedge \psi$ is $\neg(\neg\varphi \vee \neg\psi)$. In other words, a probability is in fact a function on the Boolean algebra of classes of classically equivalent formulas.

## Can the probability of a formula be understood as the truth degree of the same formula ?

Clearly not in the truth-functional case: just because probabilities are not truth-functional. However this is possible in the non-truth functional case. Let us mention for instance the paper [Gerla, 1994], where the author exhibits an abstract, non-truth functional fuzzy logical system whose set of interpretations consists of all probabilities on the set of all formulas and presents a complete deductive system for this.

## Can we understand the probability of a formula as the truth degree of another one ?

Our claim is that we can when the other formula expresses something like that the former one is "probable". This is the heart of our approach. Probability preserves classical equivalence and therefore "understands" formulas as crisp propositions. But probability is just a variable (like pressure, temperature, etc.) and we may make fuzzy assertions on it: if $\varphi$ is any formula we may say "$\varphi$ is *probable*" or "*probability-of-$\varphi$ is high*", and these are typical fuzzy propositions. Such approach was suggested in [Hajek and Harmancova, 1994]; fuzzy propositions about probabilities are also discussed in [Zimmermann, 1991]. Our aim is to describe a fuzzy theory in the frame of the truth-functional Lukasiewicz-Pavelka's logic which naturally relates to probability theory. Notice that our approach will clearly distinguish between propositions like "($\varphi$ is probable) and ($\psi$ is probable)" on the one hand and "($\varphi \wedge \psi$) is probable" in the other.

## Fuzzy theories and their logic

Following Pavelka [Pavelka, 1979], we define a *fuzzy theory* to be just a fuzzy set of formulas: if $T$ is a fuzzy theory and $T(\varphi) = x$ (denoting that the membership degree of $\varphi$ to $T$ is $x$) then $\varphi$ is an axiom to the degree $x$. Semantics is given by a set *Sem* of fuzzy sets of formulas; each element $M$ of *Sem* is understood as an interpretation of the language, i.e. $M(\varphi) = x$ is read as "$\varphi$ is true in $M$ to the degree $x$". $M$ is a model of $T$ if for any $\varphi$, $M(\varphi) \geq T(\varphi)$, i.e. each formula is at least as much true in $M$ as the $T$-degree of axiomness demands.

In a truth-functional approach, *Sem* is the set of all evaluations of formulas obtained from evaluations of propositional variables by means of some particular truth functions, e.g. the Lukasiewicz truth functions $\rightarrow, \neg$ above mentioned. But let us stress that other choices which lead to non-truth functional systems are also possible. In any case one should try to exhibit some notion of proof and try to prove some completeness result.

The paper is organized as follows. After this introduction we survey in Section 2 the Rational Pavelka's

Logic – a generalization of Lukasiewicz's logic discovered by Pavelka and simplified by Hájek. In Section 3 we present our fuzzy theory of probability and prove a completeness result. In Section 4 we comment on possible extensions and uses of the proposed approach. Finally, Section 5 contains some discussion on open problems and concluding remarks.

## 2    Rational Pavelka's Logic

Lukasiewicz's infinitely-valued logic only allows us to prove 1-tautologies, but in fuzzy logic we are interested in inference from partially true assumptions, admitting that the conclusion will also be partially true. Rational Pavelka's Logic $RPL$ is an extension of Lukasiewicz's infinitely-valued logic admitting graded formulas and graded proofs. It is described in a simple formalization in [Hájek, 1995]. Since the approach described in this paper strongly relies on this logic, here we present the main notions and properties of it.

**2.1 Formulas** are built from propositional variables $p_1, p_2, \ldots$ and truth constants $\bar{r}$ for each *rational* $r \in [0,1]$ using connectives $\rightarrow$ and $\neg$. Other connectives are defined from these ones. In particular, among others, Pavelka defines two conjunctions and two disjunctions exactly as in Lukasiewicz's logic, i.e.

| | | |
|---|---|---|
| $\varphi \,\&\, \psi$ | stands for | $\neg(\varphi \rightarrow \neg\psi)$ |
| $\varphi \,\underline{\vee}\, \psi$ | stands for | $\neg\varphi \rightarrow \psi$ |
| $\varphi \vee \psi$ | stands for | $(\varphi \rightarrow \psi) \rightarrow \psi$ |
| $\varphi \wedge \psi$ | stands for | $\neg(\neg\varphi \vee \neg\psi)$ |
| $\varphi \leftrightarrow \psi$ | stands for | $(\varphi \rightarrow \psi) \wedge (\psi \rightarrow \varphi)$ |

Taking into account the Lukasiewicz's truth functions corresponding to $\rightarrow$ and $\neg$, it is easy to check that the truth functions for the above connectives are the following ones:

$$
\begin{aligned}
r \,\&\, s &= max(\mathbf{0}, r + s - 1) \\
r \,\underline{\vee}\, s &= min(r + s, 1) \\
r \vee s &= max(r, s) \\
r \wedge s &= min(r, s) \\
r \leftrightarrow s &= min(1 - r + s, 1 - s + r)
\end{aligned}
$$

An *evaluation* of atoms is now a mapping of atomic propositions into $[0,1]$. Such mappings extend uniquely to an evaluation of all formulas respecting the above truth functions.

A *graded formula* is a pair $(\varphi, r)$ where $\varphi$ is a formula and $r \in [0,1]$ is rational. Such a formula is understood as saying that "the truth value of $\varphi$ is at least $r$".

*Logical axioms* are:

(i) axioms of Lukasiewicz's logic (all in degree 1)

$$
\begin{aligned}
\varphi &\rightarrow (\psi \rightarrow \varphi) \\
(\varphi \rightarrow \psi) &\rightarrow ((\psi \rightarrow \chi) \rightarrow (\varphi \rightarrow \chi)) \\
(\neg\varphi \rightarrow \neg\psi) &\rightarrow (\psi \rightarrow \varphi) \\
((\varphi \rightarrow \psi) \rightarrow \psi) &\rightarrow ((\psi \rightarrow \varphi) \rightarrow \varphi)
\end{aligned}
$$

(ii) bookkeeping axioms: (for arbitrary rational $r, s \in [0,1]$):

$$
\bar{r} \text{ in degree } r,
$$
$$
\overline{\neg r} \leftrightarrow \neg\bar{r} \text{ in degree 1,}
$$
$$
\overline{r \rightarrow s} \leftrightarrow (\bar{r} \rightarrow \bar{s}) \text{ in degree 1. } [1]
$$

*Deduction rules* are:

(i) *modus ponens*: from $(\varphi, r)$ and $(\varphi \rightarrow \psi, s)$ derive $(\psi, r \,\&\, s)$

(ii) *truth constant introduction*: from $(\varphi, s)$ derive $(\bar{r} \rightarrow \varphi, r \rightarrow s)$.

We define a *graded proof* from a fuzzy theory $T$ as a sequence of graded formulas

$$
(\varphi_1, r_1), \ldots, (\varphi_n, r_n)
$$

such that for each $i$, $(\varphi_i, r_i)$ is either a logical axiom (i.e. $\varphi_i$ is a logical axiom in degree $r_i$) or $(\varphi_i, r_i)$ is an axiom of $T$ (i.e. $T(\varphi_i) = r_i$) or $(\varphi_i, r_i)$ follows from some previous member(s) of the sequence by a deduction rule. We say that $T$ *proves $\varphi$ in degree $r$*, denoted $T \vdash (\varphi, r)$, if there is a graded proof from $T$ whose last element is $(\varphi, r)$. The *provability degree* of $\varphi$ in $T$ is $|\varphi|_T = \sup\{r \mid T \vdash (\varphi, r)\}$. The *truth degree* of $\varphi$ in $T$ is $\|\varphi\|_T = \inf\{e(\varphi) \mid e \text{ evaluation, } e \text{ model of } T\}$. Notice that both $\|\varphi\|_T$ and $|\varphi|_T$ may be irrational.

**2.2 Completeness theorem for RPL.** For each $T$ and $\varphi$,

$$
|\varphi|_T = \|\varphi\|_T
$$

i.e. the provability degree equals to the truth degree.

## 3    A fuzzy logic of probability

In this section we are going to define a fuzzy theory in $RPL$, that we shall call $FP$, directly related to probability theory. We start with a set of propositional variables $p, q, \ldots$ and the set of all propositional formulas built from them. Since we shall be interested in probabilities of these formulas, and hence in classical equivalence, we shall only use for them one conjunction and one disjunction, say $\wedge$ and $\vee$. We call these formulas *crisp formulas*. As suggested in [Hájek and Harmancova, 1994], we associate with each crisp formula $\varphi$ a new propositional variable $f_\varphi$, which will be read as "$\varphi$ is $PROBABLE$", or "$PROBABILITY\_OF\_\varphi$ is $HIGH$". This is understood as a fuzzy proposition, and given a probability $P$, we are free to define $e(f_\varphi) = P(\varphi)$, i.e. assign the probability value $P(\varphi)$ as the truth-value of $f_\varphi$. We may call the variables of the form $f_\varphi$ *fuzzy propositional variables* and they will be taken as the propositional variables of our fuzzy theory $FP$. Next we precisely define the $FP$ theory and show it is probabilistically meaningful.

---

[1] *Examples* of bookkeeping axioms: for $r = 0.4$ and $s = 0.3$ we get $\overline{0.6} \leftrightarrow \neg\overline{0.4}$ and $\overline{0.9} \leftrightarrow (\overline{0.4} \rightarrow \overline{0.3})$.

**3.1 Syntax** of *FP.FP-formulas* are just *RPL*-formulas built from fuzzy propositional variables, i.e. formulas built from variables of the form $f_\varphi$ using connectives. The *Axioms* of *FP* are those of *RPL* (see above) plus:

(FP1) $(f_\varphi, 1)$ for $\varphi$ being an axiom of classical propositional logic (the obvious three schemes),

(FP2) $(f_{\varphi \to \psi} \to (f_\varphi \to f_\psi), 1)$ for all $\varphi, \psi$,

(FP3) $(f_{\neg \varphi} \leftrightarrow \neg f_\varphi, 1)$ for each $\varphi$, and

(FP4) $(f_{\varphi \vee \psi} \leftrightarrow [(f_\varphi \to f_{\varphi \wedge \psi}) \to f_\psi], 1)$ for each $\varphi, \psi$.

Notice that (FP3) and (FP4) axioms are direct translations of two of the well-known axioms of probability, namely the relationship between the probability of one proposition and its negation and the finitely additivity property. Axioms (FP1) and (FP2) guarantee the preservation of classical equivalence and the monotonicity as it is proved in the next lemma and corollary.

**3.2 Lemma.** If $\varphi$ is a boolean tautology (i.e. provable in boolean propositional calculus) then *FP* proves $f_\varphi$ in degree 1.

**3.3. Corollary.** For any "crisp" formulas $\varphi$ and $\psi$ we have:

(1) If $\varphi \to \psi$ is a boolean tautology then *FP* proves $f_\varphi \to f_\psi$ in degree 1.

(2) Hence if $\varphi \leftrightarrow \psi$ is a boolean tautology then *FP* proves $f_\varphi \leftrightarrow f_\psi$ in degree 1.

(3) *FP* proves $f_{\varphi \wedge \psi} \to f_\varphi$ in degree 1.

Next we show that *FP* has exactly the intended semantics , that is, models of *FP* are defined by probability functions on the set of crisp formulas.

**3.4 Theorem.** An evaluation $e$ of atomic *FP*-formulas is a model of the theory *FP* if, and only if, the mapping $P$ defined on crisp formulas by $P(\varphi) = e(f_\varphi)$ is a finitely additive probability on crisp formulas.

*Proof.*

(1) Let $e$ be a model of *FP* and define $P(\varphi) = e(f_\varphi)$ for all $\varphi$. If $\varphi$ is classically equivalent to $\psi$ then *FP* ⊢ $(f_\varphi \leftrightarrow f_\psi, 1)$ by (2) of Corollary 3.3, hence $e(f_\varphi \leftrightarrow f_\psi) = 1$, $e(f_\varphi) = e(f_\psi)$ and therefore $P(\varphi) = P(\psi)$; thus $P$ preserves logical equivalence. Clearly, *FP* ⊢ $(f_{true}, 1)$ and *FP* ⊢ $(\neg f_{false}, 1)$, hence $P(true) = 1$ and $P(false) = 0$. Also $P(\neg \varphi) = \neg P(\varphi)$ is clear from (FP3). Now take arbitrary $\varphi, \psi$ and put $a = P(\varphi \vee \psi)$, $b = P(\varphi)$, $c = P(\psi)$, $d = P(\varphi \wedge \psi)$. By (FP4), $a = (b \to d) \to c$ and $b \geq d$ (by Corollary 3.3); thus $a = (1 - b + d) \to c$. Now $\psi \to (\varphi \to (\varphi \wedge \psi))$ is a Boolean tautology, hence *FP* ⊢ $(f_\psi \to (f_\varphi \to f_{\varphi \wedge \psi}), 1)$ and hence $c = e(f_\psi) \leq e(f_\varphi \to f_{\varphi \wedge \psi}) = 1 - b + d$; thus $a = (1 - b + d) \to c = 1 - (1 - b + d) + c =$

$b + c - d$. Thus $P$ is a probability.

(2) Conversely, assume that $P$ is a probability on crisp formulas and put $e(f_\varphi) = P(\varphi)$. We verify that $e$ assigns 1 to each axiom of *FP*. Clearly, if $\varphi$ is an axiom of classical logic then $\varphi$ is a Boolean tautology and hence $e(f_\varphi) = P(\varphi) = 1$. This verifies (FP1). To verify (FP2) we show $P(\varphi \to \psi) \leq (P(\varphi) \to P(\psi))$. Put now $P(\varphi \wedge \psi) = a$, $P(\varphi \wedge \neg \psi) = b$, $P(\neg \varphi \wedge \psi) = c$, $P(\neg \varphi \wedge \neg \psi) = d$; then $P(\varphi \to \psi) = 1 - b$, whereas $(P(\varphi) \to P(\psi)) = 1 - (a + b) + (a + c) = 1 - b + c \geq P(\varphi \to \psi)$ as desired. Under the present meaning of $a, b, c, d$ we have $e(f_{\varphi \vee \psi}) = P(\varphi \vee \psi) = a + b + c$, and $e((f_\varphi \to f_{\varphi \wedge \psi}) \to f_\psi) = ((a + b) \to a) \to (a + c) = (1 - b) \to (a + c) = a + b + c$. This verifies (FP4); (FP3) is evident. □

In [Hajek and Harmancova, 1994] the authors raised the question whether it could be possible for a fuzzy theory to have an axiomatization probabilistically complete in some sense. Here we give a positive answer.

**3.5 Definition.** A fuzzy theory $T$ is *stronger* than *FP* if for each formula $\Phi$ in the language of *FP*, $T(\Phi) \geq FP(\Phi)$ (i.e. all the axioms (FP1) ... (FP4) get the value 1 in $T$). A probability $P$ on crisp formulas is a *model* of $T$ if the corresponding evaluation $e_P$ of atoms of *FP*, defined as $e_P(f_\varphi) = P(\varphi)$, is a model of $T$.

**3.6 Corollary.** Let $T$ be a fuzzy theory stronger than *FP*. Then, for each FP-formula $\Phi$,

$|\Phi|_T = \inf\{e_P(\Phi) \mid P \text{ probability}, P \text{ model of } T\}$.

This follows directly from completeness of *RPL* and from theorem 3.4.

**3.7 Corollary. (Probabilistic Completeness for *FP*)** In particular, for each crisp formula $\varphi$,

$|f_\varphi|_T = \inf\{P(\varphi) \mid P \text{ probability}, P \text{ model of } T\}$,

$1 - |f_{\neg \varphi}|_T = \sup\{P(\varphi) \mid P \text{ probability}, P \text{ model of } T\}$.

This result tells us that if $T \vdash (f_\varphi, r)$ then for every probability $P$ which is a model of $T$, $P(\varphi) \geq r$; and also that if $T \nvdash (f_\varphi, r)$ (i.e. there is no $T$-proof of $\varphi$ to the degree $r$) then for each $r' > r$ there exists a probability $P$ which is a model of $T$ and such that $P(\varphi) < r'$.

**3.8 Remarks.** Axioms (FP1) and (FP2) could be replaced by other two (less elegant) axioms, namely by:

(FP1') $(f_{true}, 1)$, and

(FP2') $(f_\varphi \to f_\psi, 1)$, for any $\varphi$ and $\psi$ such that $\varphi \to \psi$ is a boolean tautology.

Notice that (FP2') is a direct expression of the monotonicity of probability measures with respect to set inclusion. Notice also that (FP4) may be replaced in turn by the following axiom:

$$(\text{FP4}') \quad ((f_{\varphi \vee \psi} \to f_\varphi) \to (f_\psi \to f_{\varphi \wedge \psi}), 1)$$

which is another equivalent expression for the additivity property. As a final remark, let us mention also that in $FP$ is not obvious how to represent statements about *strict* lower or upper bounds of probability.

**3.9 Example.** We present here an example of a proof in $FP$, in particular we show how to prove that $f_\varphi \leftrightarrow (f_{\varphi \wedge \psi} \veebar f_{\varphi \wedge \neg \psi})$ is a theorem of $FP$, corresponding to the well known property of probability functions stating that $P(\varphi) = P(\varphi \wedge \psi) + P(\varphi \wedge \neg \psi)$. Clearly,

$$FP \vdash (f_\varphi \leftrightarrow f_{(\varphi \wedge \psi) \vee (\varphi \wedge \neg \psi)}, 1);$$

by (FP 4), writing $\varphi 1$ for $\varphi \wedge \psi$ and $\varphi 0$ for $\varphi \wedge \neg \psi$, we get the following chain of deductions:

$$
\begin{array}{lll}
FP & \vdash & (f_{\varphi 1 \vee \varphi 0} \leftrightarrow ((f_{\varphi 1} \to f_{\varphi 1 \wedge \varphi 0}) \to f_{\varphi 0}), 1), \\
FP & \vdash & (f_{\varphi 1 \wedge \varphi 0} \leftrightarrow \overline{0}, 1), \\
FP & \vdash & ((f_{\varphi 1} \to f_{\varphi 1 \wedge \varphi 0}) \leftrightarrow \neg f_{\varphi 1}, 1), \\
FP & \vdash & (f_{\varphi 1 \vee \varphi 0} \leftrightarrow (\neg f_{\varphi 1} \to f_{\varphi 0}), 1), \\
FP & \vdash & (f_{\varphi 1 \vee \varphi 0} \leftrightarrow (f_{\varphi 1} \veebar f_{\varphi 0}), 1), \\
FP & \vdash & (f_\varphi \leftrightarrow (f_{\varphi \wedge \psi} \veebar f_{\varphi \wedge \neg \psi}), 1)
\end{array}
$$

This completes the example.

# 4   Possible extensions

The approach described so far is suitable for further developing at least along two main streams. On the one hand, we obviously need to extend $RPL$ with new connectives if we want to deal with conditional probabilities. This is addressed in subsection (a). On the other hand, the proposed approach can be easily adapted to cope with other uncertainty models, the main point being to replace the characteristic axioms of probability theory in $FP$ by the corresponding axioms characterizing other uncertainty models. As a matter of example, we shall provide in subsection (b) the fuzzy theory $FN$ corresponding to Possibility Theory and prove its completeness.

## (a) Dealing with conditional probabilities

The first idea in extending the framework presented in the previous section to deal with conditional probabilities is to look for the possibility of expressing conditional probabilitiy values as truth-values of fuzzy formulas, just as it has done in Section 3 for unconditional probabilities. Obviously, to do so, we need to introduce new connectives in the language, $\otimes$ and $\otimes{\rightarrow}$, corresponding to the product conjunction and its residuated implication (division) respectively. Then, for instance, $P(\psi \mid \varphi) \geq \alpha$ could be expressed as $(f_\varphi \otimes{\rightarrow} f_{\varphi \wedge \psi}, \alpha)$. This approach has technical problems due to the lack of continuity of the truth function associated to $\otimes{\rightarrow}$ (see section 5). However, assuming that $P(\varphi) > 0$, one can always express the inequality $P(\psi \mid \varphi) \geq \alpha$ as $P(\varphi \wedge \psi) \geq \alpha P(\varphi)$. Therefore, as a first step, we can focus on extending the Rational Pavelka's Logic with only the new conjunction connective $\otimes$, having the product as its corresponding truth function. This is done below by defining the fuzzy theory $RPL^+$, the extension of $RPL$, as follows.

**4.1 Syntax.** Formulas of $RPL^+$ are built as in $RPL$, just adding the connective $\otimes$ to the language. Logical axioms of $RPL^+$ are those of $RPL$ plus

- monotonicity:

$$(\varphi \to \psi) \to ((\varphi \otimes \chi) \to (\psi \otimes \chi))$$

$$(\varphi \to \psi) \to ((\chi \otimes \varphi) \to (\chi \otimes \psi))$$

- bookeeping:

$$\overline{r} \otimes \overline{s} \leftrightarrow \overline{r \times s}$$

all of them in degree 1.

This extension can be proved to be complete w.r.t. the above semantics, that is, the following theorem holds.

**4.2 Completeness theorem for $RPL^+$.** For each theory $T$ and formula $\varphi$ of $RPL^+$,

$$|\varphi|_T = \|\varphi\|_T,$$

i.e. in $RPL^+$ the provability degree also equals to the truth degree.

Now we are ready to define the probabilistic fuzzy theory $FP^+$ analogously to $FP$, just by replacing the axioms of $RPL$ by those of $RPL^+$. In the language of $FP^+$ we are actually able to express statements about conditional probabilities by means of formulas like

$$(\overline{\alpha} \otimes f_\varphi \to f_{\varphi \wedge \psi}, 1)$$

expressing that the conditional probability of $\psi$ given $\varphi$ is not smaller than $\alpha$, provided that the probability of $\varphi$ is known to be greater than 0. This is formalized by next theorem.

**4.3 Theorem.** Let $T$ be a theory stronger than $FP^+$ and let $\varphi$ a crisp proposition such that $|\varphi|_T > 0$. Then $|\overline{\alpha} \otimes f_\varphi \to f_{\varphi \wedge \psi}|_T = 1$ if and only if, for each probability P which is a model of $T$, it is the case that $P(\psi \mid \varphi) \geq \alpha$.

*Proof.* If $|\overline{\alpha} \otimes f_\varphi \to f_{\varphi \wedge \psi}|_T = 1$ then by completeness we have that $e_P(\overline{\alpha} \otimes f_\varphi \to f_{\varphi \wedge \psi}) = 1$ for any probability $P$ model of T, i.e. $\alpha \cdot P(\varphi) \leq P(\varphi \wedge \psi)$, and hence $P(\psi \mid \varphi) \geq \alpha$, provided that $P(\varphi) > 0$, but this is guaranteed by having $|\varphi|_T > 0.\square$

Moreover, statements about conditional independence saying that for instance $\varphi$ and $\psi$ are independent given $\chi$ could be also expressed by means of axioms extending $FP^+$ as

$$((f_{\varphi \wedge \psi \wedge \chi} \otimes f_\chi) \leftrightarrow (f_{\psi \wedge \chi} \otimes f_{\varphi \wedge \chi}), 1)$$

$$((f_{\varphi \wedge \psi \wedge \neg \chi} \otimes f_{\neg \chi}) \leftrightarrow (f_{\psi \wedge \neg \chi} \otimes f_{\varphi \wedge \neg \chi}), 1)$$

Further results about the probabilistic completeness of $FP^+$ will deserve future attention.

## (b) A fuzzy logic for possibility theory

As an example of the fact that fuzzy logic is also a suitable framework to describe other uncertainty models different from probability theory, we present below the fuzzy theory $FPS$ to reason with formulas valued with possibility and necessity degrees. Possibility theory, as uncertainty model, has been widely developed from a logical point of view under the so-called *Possibilistic Logic* (see e.g. [Dubois *et al.*, 1994] for an extensive survey). Possibilistic logic obviously does not need the whole machinery we are going to use, but nevertheless we still think it can be interesting for exemplifying purposes. Thus, now we are interested in associating to each crisp formula $\varphi$ a fuzzy formula $f_\varphi$, which will be read as "$\varphi$ is *NECESSARY*" or "$\varphi$ is *CERTAIN*", in such a way that the truth-degree of $f_\varphi$ represents the necessity degree (in the sense of necessity measures) of $\varphi$, and therefore the truth degree of $\neg f_{\neg \varphi}$ represents the possibility degree of $\varphi$.

### 4.4 Syntax of $FPS$. $FPS$-formulas are just $FP$-formulas, i.e. formulas built from fuzzy propositional variables of the form $f_\varphi$ using connnectives. *Axioms* of $FPS$ are those of $RPL$ (see section 3) plus:

(FPS1) $(f_\varphi, 1)$ for $\varphi$ being an axiom of classical propositional logic, ( = (FP1))

(FPS2) $(f_{\varphi \to \psi} \to (f_\varphi \to f_\psi), 1)$ for all $\varphi, \psi$, ( = (FP2))

(FPS3) $(\neg f_{false}, 1)$, and

(FPS4) $((f_\varphi \wedge f_\psi) \leftrightarrow f_{\varphi \wedge \psi}, 1)$.

Notice that, if we denote $\neg f_{\neg \varphi}$ by $g_\varphi$, we would get dual axioms corresponding to possibility measures, in particular (FPS1) and (FPS3) are also valid for propositional variables of type $g_\varphi$, and (FPS4) would be equivalently expressed as

(FPS4') $((g_\varphi \vee g_\psi) \leftrightarrow g_{\varphi \vee \psi}, 1)$

*Caution:* Note that the obvious analogon of (FPS2) for possibilities is *not* sound.

Analogous results to those for $FP$ can be proved for $FPS$.

### 4.5 Lemma. For any "crisp" formulas $\varphi$ and $\psi$ we have:

(1) If $\varphi$ is a boolean tautology (i.e. provable in boolean propositional calculus) then $FPS$ proves $f_\varphi$ in degree 1.

(2) If $\varphi$ is a boolean antitautology (i.e. $\neg \varphi$ is provable in boolean propositional calculus) then $FPS$ proves $\neg f_\varphi$ in degree 1.

(3) If $\varphi \to \psi$ is a boolean tautology then $FPS$ proves $f_\varphi \to f_\psi$ in degree 1.

(4) If $\varphi \leftrightarrow \psi$ is a boolean tautology then $FPS$ proves $f_\varphi \leftrightarrow f_\psi$ in degree 1.

### 4.6 Theorem. An evaluation $e$ of atomic $FPS$-formulas is a model of the theory $FPS$ iff the mapping $N$ defined on crisp formulas by $N(\varphi) = e(f_\varphi)$ is a necessity measure on crisp formulas, i.e. $N(True) = 1, N(False) = 0$ and $N(\varphi \wedge \psi) = min(N(\varphi), N(\psi))$.

*Proof.*

We only prove that if $N$ is a necessity on crisp formulas then the evaluation defined as $e(f_\varphi) = N(\varphi)$ assigns 1 to axiom (FPS 2). The rest is straightforward. Thus, we have to prove that $Nec(\varphi \to \psi) \leq (Nec(\varphi) \to Nec(\psi))$. By cases:

- if $Nec(\varphi) \leq Nec(\psi)$ it is trivial.

- suppose then that $Nec(\varphi) > Nec(\psi)$. Since $Nec(\psi) = min(Nec(\psi \vee \varphi), Nec(\psi \vee \neg \varphi))$ and $Nec(\psi \vee \varphi) \geq Nec(\varphi) > Nec(\psi)$ the only possibility is that $Nec(\psi) = Nec(\psi \vee \neg \varphi)$, and therefore $(Nec(\varphi) \to Nec(\psi)) \geq Nec(\psi) = Nec(\psi \vee \neg \varphi)$, which ends the proof.□

### 4.7 Definition. A fuzzy theory $T$ is *stronger* than $FPS$ if for each formula $\Phi$ in the language of $FPS$, $T(\Phi) \geq FPS(\Phi)$ (i.e. all the axioms (FPS 1); ... (FPS 4) get the value 1 in $T$). A necessity function $N$ on crisp formulas is a *model* of $T$ if the corresponding evaluation $e_N$ of atoms of $FPS$ defined as $e_N(f_\varphi) = N(\varphi)$ is a model of $T$.

The completeness result for $FPS$, analogous again to that for $FP$ is given in the following theorem.

### 4.8 Theorem. Let $T$ be a fuzzy theory stronger than $FPS$. Then, for each $FPS$-formula $\Phi$,

$$\mid \Phi \mid_T = \inf\{e_N(\Phi) \mid N \text{ necessity, } N \text{ model of } T\}.$$

In particular, for each crisp formula $\varphi$ we have:

$$\mid f_\varphi \mid_T = \inf\{N(\varphi) \mid N \text{ necessity, } N \text{ model of } T\}.$$

Notice that the well-known possibilistic resolution principle for necessity valued clauses, saying that from

"$N(\varphi \vee \psi) \geq \alpha_1$" and "$N(\neg \varphi \vee \chi) \geq \alpha_2$"

it can be inferrred

"$N(\psi \vee \chi) \geq \alpha_1 \wedge \alpha_2$",

is now a derivable inference rule in $FPS$. Namely, since $[(\varphi \vee \psi) \wedge (\neg \varphi \vee \chi)] \to (\psi \vee \chi)$ is a boolean tautology, by lemma 4.5(3), $FPS$ proves $f_{(\varphi \vee \psi) \wedge (\neg \varphi \vee \chi)} \to f_{\psi \vee \chi}$ with degree 1. By (FPS4), $FP$ proves also $(f_{\varphi \vee \psi} \wedge f_{\neg \varphi \vee \chi}) \to f_{\psi \vee \chi}$ with degree 1. Now the proof easily comes by modus ponens taking into account that the completeness of $RPL$ allows us to infer $(f_{\varphi \vee \psi} \wedge f_{\neg \varphi \vee \chi}, \alpha_1 \wedge \alpha_2)$ from $(f_{\varphi \vee \psi}, \alpha_1)$ and $(f_{\neg \varphi \vee \chi}, \alpha_2)$.

## 5  Conclusions and open problems

In this paper we have been concerned about stressing the conceptual differences between fuzzy logic and probability, and we have shown, as a main result, that both notions can be consistently used together to define a fuzzy theory $FP$ in the Rational Pavelka's Logic (an extension of Lukasiewicz's logic with truth constants and graded proofs) which is closely related to probability theory. The basic approach has been: the probability of a crisp formula $\varphi$ is understood as the truth degree of the fuzzy atomic proposition $f_\varphi$ saying that "$\varphi$ is probable". Models of $FP$ are in one-to-one relation to probabilities on the set of crisp formulas; graded proofs of $f_\varphi$ in a fuzzy theory $T$ containing $FP$ give lower (and upper) bounds of $P(\varphi)$ for all probabilities $P$ that are models of $T$. This is hoped to contribute to the understanding of the relation of fuzzy logic and probability. Moreover we have also sketched two interesting extensions of this approach. In the first one we show the possibility of dealing with conditional probabilities inside the same framework by extending Rational Pavelka's Logic with the product conjunction connective. In the second one we have shown the possibility of adapting the proposed approach to cope with other uncertainty calculi, in particular this has been done for Possibility theory. Remaining issues to be addressed are, among others:

- a more elegant way of representing conditional probabilities by means of the product residuated implication and try to solve the problems related to the fact that this implication is not continuous and hence does not admit a Pavelka-style completeness theorem;

- axiomatization of a fuzzy theory related to belief functions. To this respect, it seems suitable to introduce in the language some modalities if we want to avoid having very cumbersome axioms corresponding to the sub-additivity properties of belief functions.

### Acknowledgements

This research has been partially supported by the COPERNICUS project MUM (10053) from the European Union. Petr Hájek has been also partially supported by the grant no. 130108 of the Academy of Science of the Czech Republic.